\documentclass[conference]{IEEEtran}
\IEEEoverridecommandlockouts

\usepackage{graphicx}
\usepackage{subfigure}
\usepackage{booktabs} 
\usepackage[T1]{fontenc}
\usepackage[ruled]{algorithm2e} 

\usepackage{flushend}
\usepackage{amsmath,amssymb,amsfonts}

\def\BibTeX{{\rm B\kern-.05em{\sc i\kern-.025em b}\kern-.08em
    T\kern-.1667em\lower.7ex\hbox{E}\kern-.125emX}}
\begin{document}

\title{FedDMF: Privacy-Preserving User Attribute Prediction using Deep Matrix Factorization}


\author{\IEEEauthorblockN{Ming Cheung}
\IEEEauthorblockA{\textit{Beta Labs, The Lane Crawford Joyce Group} \\
Hong Kong, China \\
mingcheung@lcjgroup.com}
}



\maketitle

\begin{abstract}
User attribute prediction is a crucial task in various industries. 
However, sharing user data across different organizations faces challenges due to privacy concerns and legal requirements regarding personally identifiable information. 
Regulations such as the General Data Protection Regulation (GDPR) in the European Union and the Personal Information Protection Law of the People's Republic of China impose restrictions on data sharing.
To address the need for utilizing features from multiple clients while adhering to legal requirements, federated learning algorithms have been proposed. 
These algorithms aim to predict user attributes without directly sharing the data. 
However, existing approaches typically rely on matching users across companies, which can result in dishonest partners discovering user lists or the inability to utilize all available features.
In this paper, we propose a novel algorithm for predicting user attributes without requiring user matching. Our approach involves training deep matrix factorization models on different clients and sharing only the item vectors. This allows us to predict user attributes without sharing the user vectors themselves.
The algorithm is evaluated using the publicly available MovieLens dataset and demonstrate that it achieves similar performance to the FedAvg algorithm, reaching 96\% of a single model's accuracy. 
The proposed algorithm is particularly well-suited for improving customer targeting and enhancing the overall customer experience.
This paper presents a valuable contribution to the field of user attribute prediction by offering a novel algorithm that addresses some of the most pressing privacy concerns in this area.
\end{abstract}

\section{Introduction}
\begin{figure*}
\centering
\includegraphics[width=5.5in]{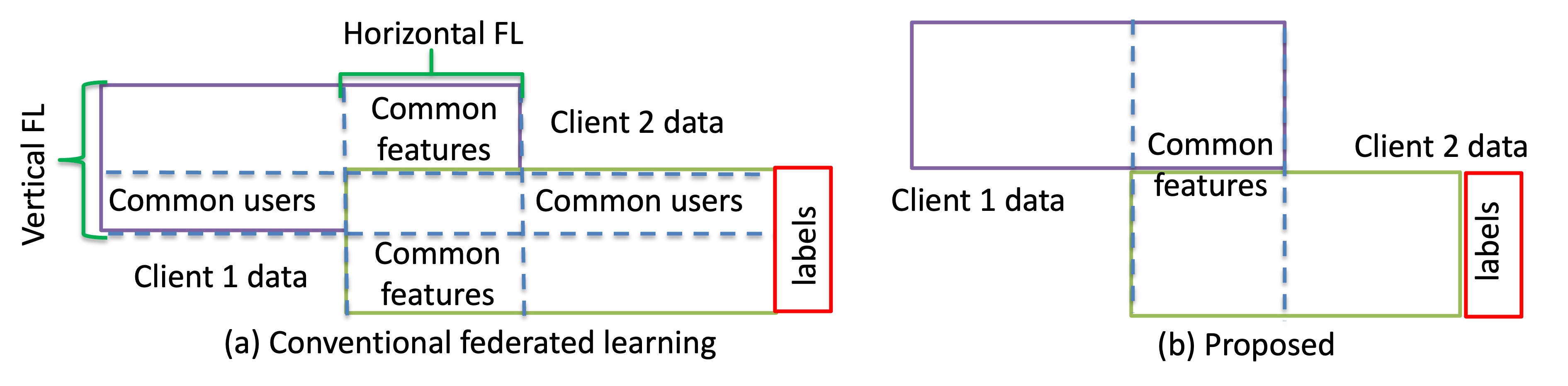}
\caption{Conventional federated learning and the proposed.}
\label{fig:Federated_learning}
\end{figure*}

User attribute prediction plays a crucial role in various industries, including retail, as it enables improved customer targeting and enhances the overall customer experience. 
However, sharing user data across different organizations poses challenges due to privacy concerns and legal requirements regarding Personal Identifiable Information (PII).
The General Data Protection Regulation (GDPR) is a European Union regulation that safeguards personal data of individuals within the EU and imposes requirements on organizations worldwide that process personal data. Similarly, the Personal Information Protection Law (PIPL) is a Chinese law focused on protecting personal information within China's territory, applicable to both domestic and foreign organizations processing personal information of Chinese residents. 
Consequently, sharing data among companies and organizations is becoming increasingly difficult.
\\
\indent
To address these challenges, federated learning algorithms have been proposed to train models without directly sharing data. 
However, previous algorithms often require matching users across companies or fail to utilize all available features. 
Fig. \ref{fig:Federated_learning} illustrates an example of federated learning for two companies.
In Fig. \ref{fig:Federated_learning} (a), two companies have distinct sets of users and features, with some common users and features existing in both clients. 
Although Company 2 possesses labels that Client 1 requires, they cannot be shared due to legal requirements, and vice versa. 
These labels represent user attributes such as gender and transaction records indicating whether a user has made a purchase. 
Horizontal federated learning focuses on extracting common features with all users for training a model \cite{gao2019privacy}. 
As a result, all users have similar feature vectors but it does not utilize all features for effective learning. 
On the other hand, vertical federated learning extracts common users with all features for model training \cite{liu2021fedct}. 
These users possess a complete set of features across both companies. 
However, the two clients must match users before training, which can pose privacy risks.
\\
\indent
To overcome these limitations, this paper proposes a novel algorithm that utilizes all features without requiring user matching. Instead of training a model with a subset of users or features, a deep matrix factorization (DMF) model is trained on different clients, and the feature vectors are shared to predict user attributes \cite{xue2017deep}. 
Such common features can be easily found from e-commerce websites without privacy concerns. 
By employing this algorithm, accurate predictions can be achieved without the need to share user vectors, thereby preserving user privacy. 
To demonstrate the effectiveness of the proposed algorithm, experiments were conducted on the publicly available MovieLens dataset. 
The results indicate that the proposed algorithm performs similarly to the FedAvg approach and achieves 96\% of a single model's performance. 
This algorithm is particularly well-suited for the retail industry, where it can enhance customer targeting and improve the overall customer experience.
This paper contributes in the following ways:

\begin{itemize}
\item Proposes a new algorithm called FedDMF, based on DMF, that eliminates the need for sharing and matching users.
\item Conducts experiments to demonstrate the effectiveness of the proposed algorithm.
\end{itemize}

The rest of the paper is structured as follows: Section~\ref{sec:related_works} provides a brief overview of related work, Section~\ref{sec:algorithm} describes the proposed algorithm in detail, Section~\ref{sec:Experimental} presents the experimental setup and results, and Section~\ref{sec:conclusion} concludes the paper and discusses future work.

\section{Related Works}
\label{sec:related_works}
\noindent
Federated learning (FL) has emerged as a promising algorithm for learning from decentralized data sources while preserving user privacy \cite{shokri2015privacy}. 
There are applications in various domains, including healthcare \cite{xu2021federated, grama2020robust}, finance \cite{long2020federated}, and the retail industry \cite{flanagan2021federated, ammad2019federated}, for tasks such as recommendations, attribution, and user segmentation. FL encompasses two commonly used settings: vertical FL and horizontal FL.
In horizontal FL \cite{gao2019privacy}, the data is partitioned horizontally, with each party holding a subset of the features for all users. 
However, in this setting, the common features are limited to a small set, as depicted in Fig. \ref{fig:Federated_learning} (a). 
If there are more than two clients involved, the number of features may be insufficient to train meaningful models.
On the other hand, vertical FL \cite{liu2021fedct, fu2021vf2boost, xu2021fedv} involves partitioning the data vertically, whereby each party holds a subset of the samples. 
While vertical FL can utilize all features, it requires matching users across different clients, and there may be a limited subset of common users.
\\
\indent
While FL aims to preserve user privacy by keeping data decentralized, using FL still poses privacy risks. 
During the training process, information leakage can occur, enabling adversaries to reconstruct sensitive data or infer private information about individuals \cite{ye2022enhanced}. 
Adversaries with access to the model updates or aggregated data may attempt to extract sensitive information \cite{zhu2019deep, zhao2020idlg}.
To address these concerns, algorithms have been developed \cite{fan2020rethinking}. 
However, previous federated learning algorithms for predicting user attributes often required matching users across different companies, leading to time-consuming processes and potential violations of user privacy. 
Even using private set intersection may introduce potential information leakage and computational overhead \cite{yang2019comprehensive}, as the user identity can be recovered during the matching process.
\\
\indent
Recently, there has been growing interest in using deep matrix factorization (DMF) models for user attribute prediction \cite{de2021survey}. 
These models have proven effective in capturing latent factors that influence user preferences and modeling user-feature interactions. 
In this paper, a novel algorithm is proposed to predict user attributes across different clients without the need for user matching, while utilizing all features across different clients. 
This approach involves training a deep MF model on different clients and sharing only the feature vectors.

\begin{figure*}
\centering
\includegraphics[width=5.5in]{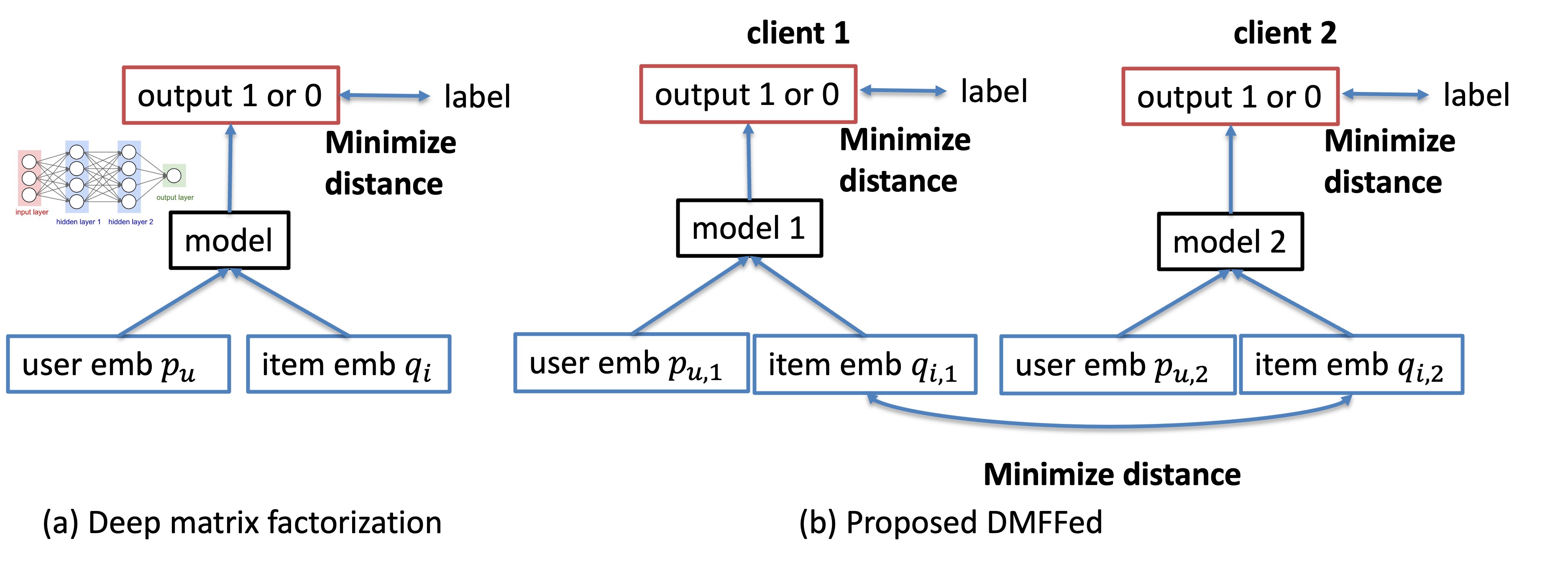}
\caption{Proposed algorithm: FedDMF.}
\label{fig:Proposed}
\end{figure*}

\section{Proposed Algorithm}
\label{sec:algorithm}
\noindent
This section presents a novel algorithm for model training that eliminates the need to match users across datasets of different clients, thereby avoiding data leakage. The proposed algorithm is introduced after providing an overview of deep matrix factorization (DMF) and federated learning (FL).

\subsection{Deep matrix factorization} 
\label{ssub:deep_cf}
\noindent
Fig. \ref{fig:Proposed} illustrates the ideas behind the proposed algorithm.
DMF, as shown in Fig. \ref{fig:Proposed}(a), is a technique that combines the power of deep learning and matrix factorization to learn low-dimensional representations of high-dimensional data \cite{xue2017deep}. 
It employs neural networks to model complex relationships and factorize the data into latent factors, facilitating efficient representation learning and recommendation systems.
In a recommendation scenario, the items are considered as features. 
The model outputs depend on whether a user has interacted with an item, with an interaction represented by a value of 1 and no interaction represented by a value of 0. 
The objective function to be minimized can be defined as follows:
\begin{equation}
\label{eq:dmf_predict}
\hat{y}_{ui} = f(\mathbf{p}_u, \mathbf{q}_i, \mathbf{W})
\end{equation}
where $\hat{y}_{ui}$ represents the predicted output for user $u$ and item $i$. 
The symbols $\mathbf{p}_u$ and $\mathbf{q}_i$ denote the learnable embedding vectors for user $u$ and item $i$, respectively. 
The function $f(\cdot)$ represents the deep learning model that takes the user embedding and item embedding as inputs and predicts the output, with $\mathbf{W}$ is the learnable weight of the deep learning model.
The model can be trained using the mean squared error (MSE), which measures the average squared difference between the predicted ratings and the actual ratings in the training data:
\begin{equation}
\label{eq:dmf_mse}
L_{MSE} = \frac{1}{N} \sum_{(u, i) \in \text{Training Data}} (\hat{y}{u,i} - y{u,i})^2
\end{equation}
where $N$ is the total number of training samples, and $y_{u,i}$ is the actual value for user $u$ and item $i$ in the training data. The model is updated using the computed gradient $\theta$ through approaches such as stochastic gradient descent \cite{ketkar2017stochastic}.

\subsection{Federated Learning} 
\label{sub:federated_learning}
\noindent
In the federated learning (FL) setting, each client trains a separate model using its local dataset, as data cannot be shared among different clients. 
The loss is calculated separately on each client using its own data. 
Once the loss is computed, the losses from all clients are combined, and the models on each client are updated accordingly.
The local models are then aggregated to obtain a global model that captures knowledge from all clients while keeping sensitive data within each party, addressing privacy concerns. 
The aggregation of local models can be performed using various techniques, such as FedAvg. 
Mathematically, the federated averaging algorithm can be expressed as follows:
\begin{equation}
\label{eq:fedavg}
\theta_{l+1} = \frac{1}{N_p}\sum_{p=1}^{N_p} \theta_p
\end{equation}
where, $\theta_{l+1}$ represents the gradient at iteration $l+1$, $\theta_p$ denotes the local model of party $p$, and $N_p$ is the total number of clients participating in the federated learning process. 
The local models can be updated using gradients.
However, as discussed earlier, both horizontal and vertical FL suffer from limitations such as a limited number of features or users, as well as privacy risks. 
In the next section, the proposed FedDMF algorithm will be introduced to address these challenges.

\begin{figure*}
\centering
\includegraphics[width=5.5in]{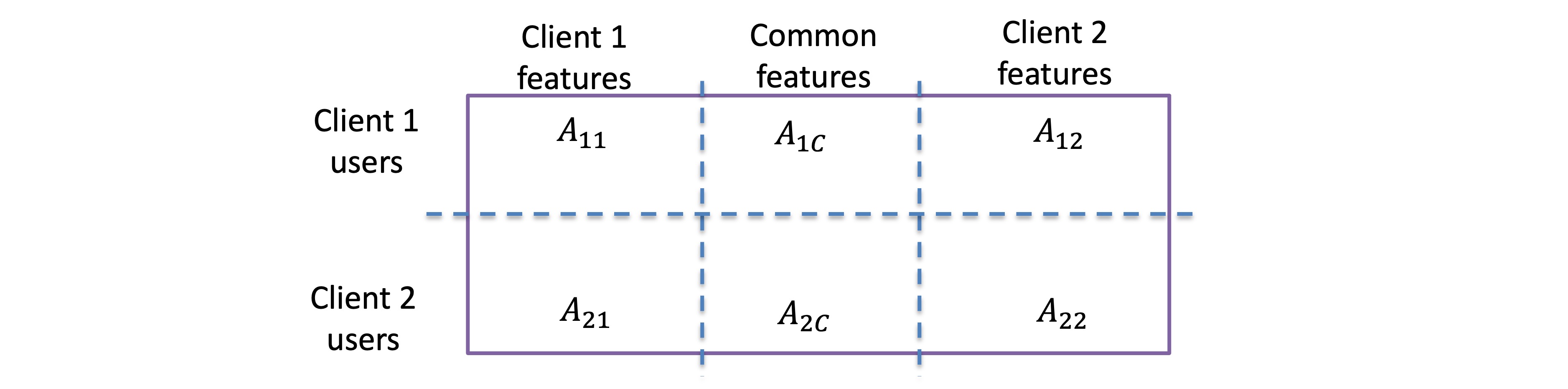}
\caption{Experimental Settings.}
\label{fig:example_exp_settings}
\end{figure*}

\begin{figure*}
\centering
\includegraphics[width=7.05in]{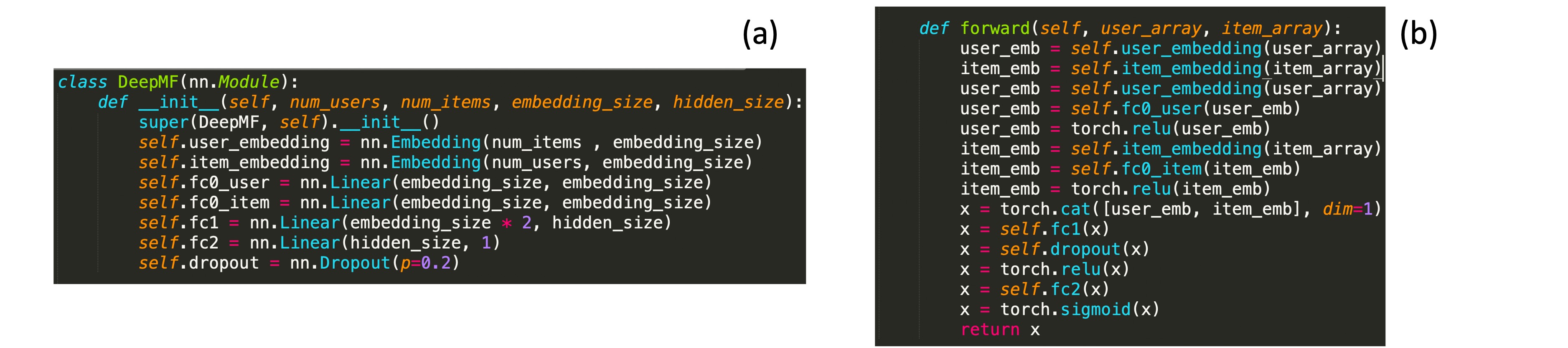}
\caption{Model structure, (a) initialization, (b) model structure.}
\label{fig:Model_structure}
\end{figure*}

\subsection{Proposed FedDMF}
\noindent
This section introduces the proposed FedDMF algorithm, which is based on DMF. 
Unlike conventional algorithms that require clients to match the identity of users across different clients, the proposed algorithm allows each client to have a separate model with non-shared parameters, minimizing the risk of data sharing.
While the user vectors may have different inputs and shapes across clients, the feature vectors must have the same shape across all clients. 
It is assumed that the features among different clients can be matched, and the feature vectors can be shared. 
This assumption is valid in scenarios such as e-commerce websites, where all websites share the same features and can be easily matched across platforms.
\\
\indent
Before training, the features are matched, and all clients have a copy of the feature matching. 
Each client then initializes a separate model with the agreed feature vector size. 
During training, each model is trained using the respective client's data.
In addition to the mean squared error (MSE) loss in Eq. \ref{eq:dmf_mse}, there is an additional loss term that measures the cosine distance between vectors of the same feature across different clients. 
It is the embedding similarity loss:

\begin{equation}
\label{eq:dmf_mse_proposed}
{L_{emb}} =  \frac{\beta}{|\textbf{P}|}.\sum_{(i_{p_1},i_{p_2}) \in \textbf{P}} max(1-cos(\mathbf{q}_{i_{p_1}}, \mathbf{q}_{i_{p_2}}),m)
\end{equation}

The modified loss function is as follows:
\begin{equation}
\label{eq:dmf_mse_proposed}
{L_a} =L_{MSE} +{L_{emb}} 
\end{equation}
where $\textbf{P}$ represents the set of feature pairs that exist in both clients. 
The new loss term is the sum of the cosine distances between all pairs ($i_{p_1}$, $i_{p_2}$) that belong to the set $\textbf{P}$. 
For each pair ($i_{p_1}$, $i_{p_2}$) in $\textbf{P}$, the loss is the maximum value between the cosine distance and $m$. 
The parameter $\beta$ is the weight of the cosine distance and prevents the pairs from being overfit, thus affecting the performance. 
Finally, the sum is divided by the size of the set $\textbf{P}$ using the fraction.
\\
\noindent
This additional loss term helps to project the feature vectors among different clients onto the same hyperspace, enabling the prediction of a vector from another model even if the model has not seen the vector before. 
The proposed algorithm allows the training of a joint model without the need to match a set of users from different clients. 
This is an improvement over previous algorithms, as the proposed FedDMF does not require matching users, utilizes all features, and does not require sharing model parameters.
In the retail industry, the features could be user-feature transaction records. 
Although the transaction records are kept private, the feature matching can be easily obtained from platforms such as e-commerce websites. The feature vectors can then be shared between clients during training.
The next section introduces the experimental setup to evaluate the performance of the proposed FedDMF algorithm.

\section{Experimental Results}
\label{sec:Experimental}
\noindent  
This section presents the experimental results of the proposed FedDMF algorithm. 
The dataset used in the experiments is first introduced, followed by the experimental settings. 

\subsection{Dataset}
\noindent
The movielens dataset, specifically the ml-latest-small version, was used in our experiments. 
This public dataset contains a subset of movie ratings by users\footnote{http://files.grouplens.org/datasets/movielens/ml-latest-small.zip}. 
The dataset was created by randomly selecting a subset of ratings from the original movielens dataset. 
It includes ratings for over 9,000 movies by more than 600 users, where each movie serves as a feature in the experiments.

\subsection{Settings}
\noindent
To simulate the Federated Learning (FL) training scenario, the dataset is split into two parts, representing two clients. 
The experimental settings are illustrated in Fig. \ref{fig:example_exp_settings}. 
The features are divided based on the number of features each client has, and a certain percentage of features are selected as common features to be used in training. 
The data is further split into different parts: client 1 and client 2. 
While there are some common features between the two clients, there are no common users.
For example, in Fig. \ref{fig:example_exp_settings}, $A_{1,1}$, $A_{1,C}$, and $A_{1,2}$ represent the data parts for user 1. 
$A_{1,1}$ and $A_{1,C}$ are used by client 1 for training, while $A_{1,2}$ serves as the testing set for client 1. 
The difference between $A_{1,1}$ and $A_{1,C}$ is that $A_{1,C}$ consists of the shared features with client 2, while $A_{1,1}$ includes the features not present in the training set of client 2. 
The features of $A_{1,2}$ belong to client 2.
During the training process, the prediction does not include the common features, but rather the features from the other client. 
For example, client 1 will predict the user-feature pair $A_{1,2}$, while client 2 will predict the user-feature pair $A_{2,1}$. 
The proposed algorithm trains the model using a classification approach, where a user who has rated a movie is classified as 1 for that user-movie pair and 0 otherwise.
\\
\indent
The DeepMF model structure, which is a neural network-based model, is depicted in Fig. \ref{fig:Model_structure}. 
The model takes as input the number of users, number of features, embedding size, and hidden size. 
The inputs to the model are the user embedding, $\mathbf{p}_u$, and the feature embedding, $\mathbf{q}_i$. 
These inputs are passed through the user and feature fully connected (FC) layers, followed by the ReLU activation function. 
Dropout is applied to the output of the FC1 layer. 
The result is then passed through the FC2 layer, followed by the sigmoid activation function, to obtain the final output, which represents the predicted value, $\hat{y}_{u,i}$. 
The embedding size and hidden size are set to 32 and 64, respectively. The batch size is 32 with a learning rate of 0.001. Each training process consists of 30 epochs, with the value of $m$ set to 0.2 and $\beta$ set to 100.
Once the model is trained, it is used to predict the features of other clients. If the predicted value is greater than 1, it is classified as 1; otherwise, it is classified as 0. The experiments are repeated three times with different training and testing sets, and the mean result is recorded.
\begin{figure*}
\centering
\includegraphics[width=7.05in]{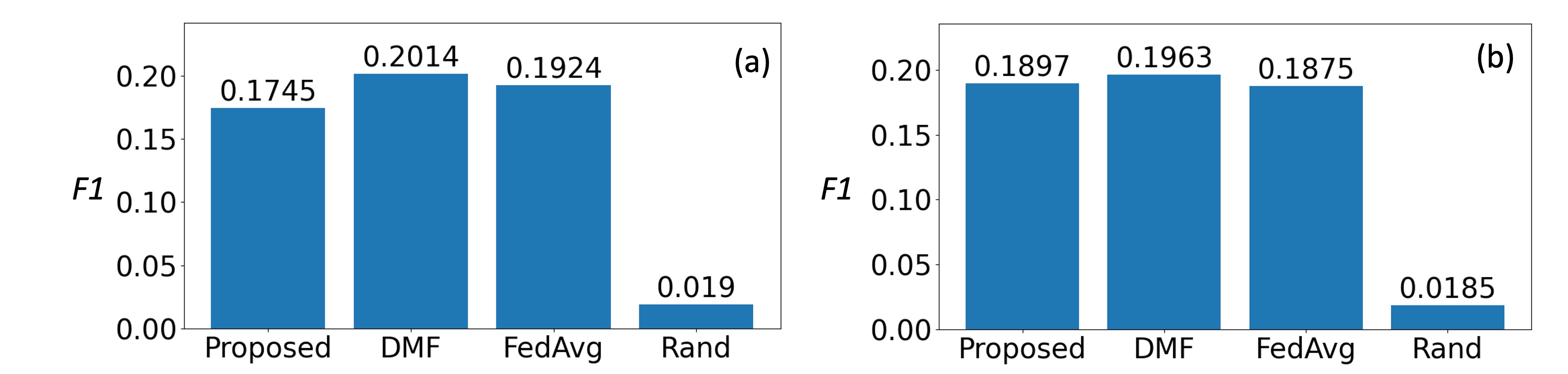}
\caption{Experimental Results with baselines: (a) training set, (b) testing set.}
\label{fig:Result_compare}
\end{figure*}

\begin{figure*}
\centering
\includegraphics[width=7.05in]{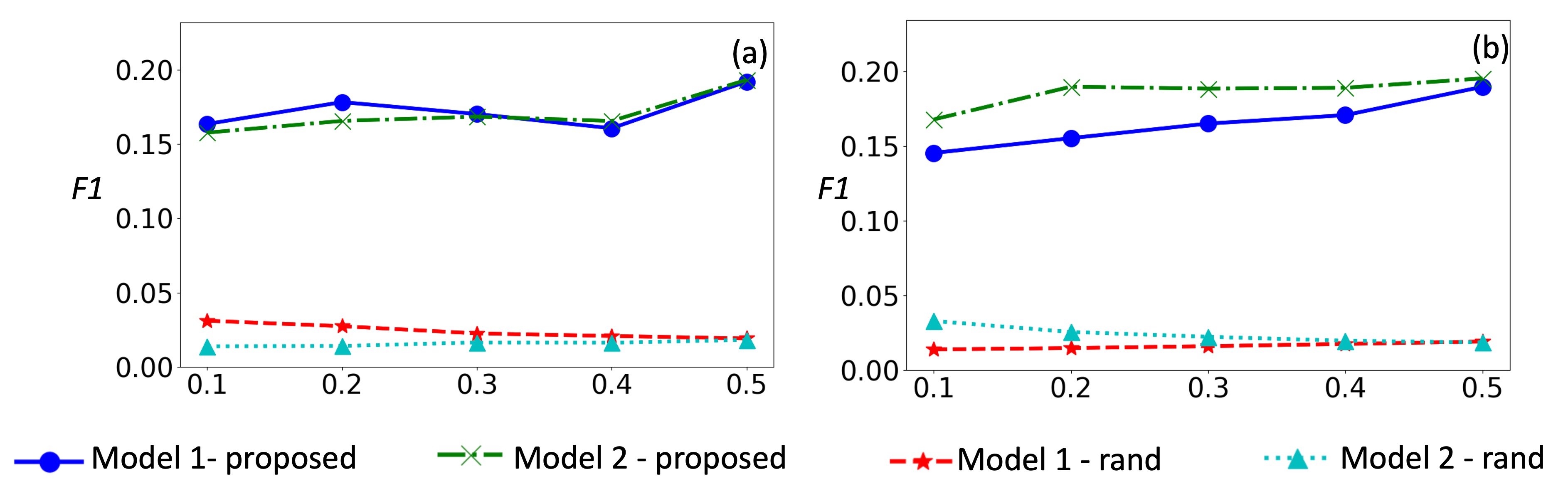}
\caption{Experimental Results with different percentages of features that client 1 has: (a) training set, (b) testing set.}
\label{fig:res_hk_feature_prop}
\end{figure*}

\subsection{Results}
\noindent
This section presents the experimental results of the proposed algorithm and compares its effectiveness with two baselines. 
The first baseline is DMF, which represents the conventional algorithm when all data is accessible and serves as the upper bound. 
The second baseline is FedAvg, where the gradient is averaged in each epoch to train the same model for both clients. 
Additionally, a random approach (rand) is included as another baseline.
The results are illustrated in Fig. \ref{fig:Result_compare}. 
In the experiments, 50\% of the features were used as common features, and each client had 50\% of the remaining features and 50\% of the users. Fig. \ref{fig:Result_compare}(a) and (b) show the training and testing $F1$ scores, respectively.
The results indicate that DMF achieves the best performance, followed by FedAvg as the second best. 
The proposed algorithm demonstrates similar performance to the FedAvg algorithm, achieving 96\% of the performance achieved by a single model. These results confirm that the proposed algorithm is capable of training a high-quality model.
In the subsequent sections, The performance of the proposed algorithm with different percentages of common features will be investigated. 
This analysis aims to provide insights into the algorithm's performance under varying conditions.

\subsection{Results of proportion of users}
\noindent
This experiment investigates how the proportion of users affects the training results. 
Similar to the previous experiments, the results are measured in terms of F1 score, with 50\% of common features and 50\% of users from client 1. The results are shown in Fig. \ref{fig:res_hk_feature_prop}.
In Fig. \ref{fig:res_hk_feature_prop}(a) and (b), the predictions of the features in the training and testing sets are presented, respectively. 
The x-axis represents the percentage of features that client 1 has, ranging from 0.1 to 0.5.
The blue curve represents the results for the first client, while the green curve represents the results for the second client.
From Fig. \ref{fig:res_hk_feature_prop}(a), it is observed that when the proportion of a client is small, the F1 score is high. 
This can be attributed to the fact that with fewer user-feature pairs to predict, the model can allocate more resources to accurately predict those pairs.
Moving to Fig. \ref{fig:res_hk_feature_prop}(b), it can be seen that the algorithm achieves the best results when both clients have an equal number of users. 
This suggests that having an equal distribution of users provides a balanced and representative training set, leading to improved prediction performance.

\begin{figure*}
\centering
\includegraphics[width=7.05in]{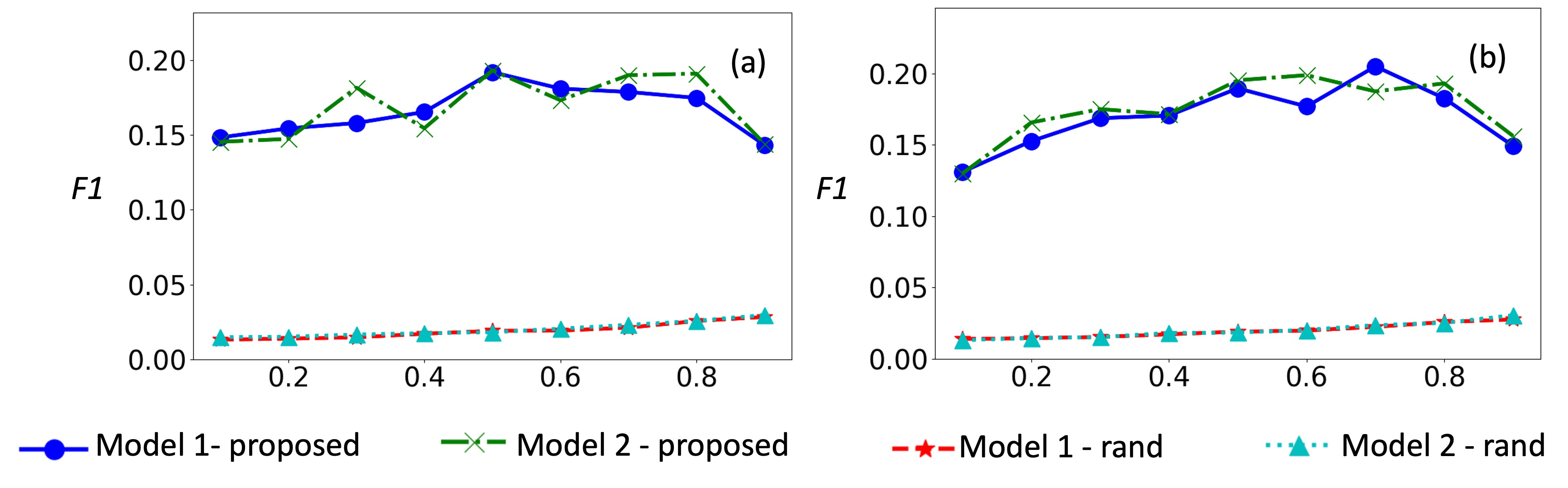}
\caption{Experimental Results with different percentage of common features: (a) training set, (b) testing set.}
\label{fig:res_com_feature_prop}
\end{figure*}

\begin{figure*}
\centering
\includegraphics[width=7.05in]{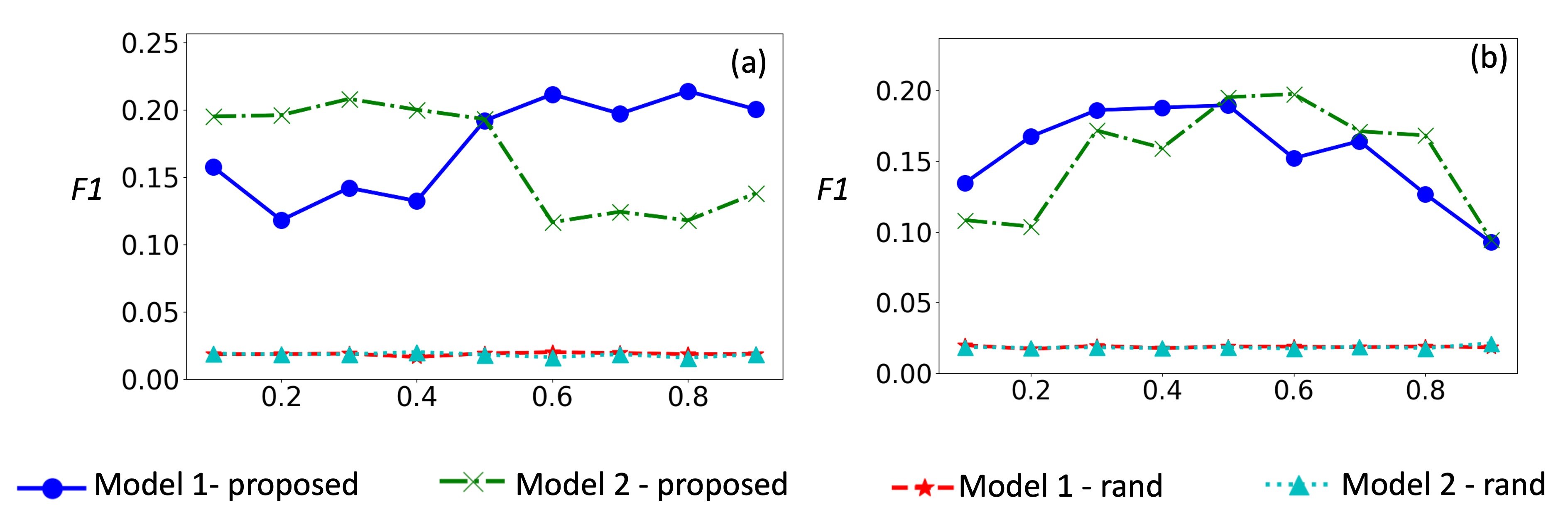}
\caption{Experimental Results with different percentage of client 1 users: (a) training set, (b) testing set.}
\label{fig:res_hk_prop}
\end{figure*}

\subsection{Common Features}
The percentage of common features plays a significant role in model training. 
An experiment was conducted to investigate how the percentage of common features affects the model's performance, and the results are presented in Fig. \ref{fig:res_com_feature_prop}.
It is important to note that in the prediction of both the training and testing sets, the common features are not considered. 
The focus is on evaluating the impact of the percentage of common features on the performance of predicting features from each client and features from the other client.
The results show that the $F1$ score initially increases with the percentage of common features for both types of features. 
This can be attributed to the fact that having more common features implies a higher likelihood of better training. 
As more features are available for training, the model can learn more effectively and improve its prediction performance.
However, beyond a certain threshold, the $F1$ score starts to drop. 
One possible reason for this decline is that, as the prediction does not consider common features, there are only a few features available in the ground truth for evaluation. This scarcity of features can lead to inaccurate predictions and lower $F1$ scores.
Another notable result is that the Rand index increases with the percentage of common features. 
This outcome can be explained by the fact that the prediction does not consider common features. 
As a result, there are fewer features to be predicted, as well as a reduced number of user-feature pairs in the ground truth. Consequently, the prediction has a higher recall and the same precision, resulting in a higher F1 score.

\subsection{Results of \% of client 1 features}
\noindent
This section investigates how the proportion of features in client 1 affects the model's performance. 
It is expected that when a client contains more users, the training results for that client should be better. 
To validate this hypothesis, an experiment was conducted, and the results are presented in Fig. \ref{fig:res_hk_prop}.
In Fig. \ref{fig:res_hk_prop}(a), the training $F1$ score reaches a high value at and after 0.5, indicating that when a client has 50\% of the users, it achieves a higher $F1$ score. 
However, when a client has less than 50\% of the users, its $F1$ score is lower. The performance for the training features remains consistent across different proportions.
The testing $F1$ score is shown in Fig. \ref{fig:res_hk_prop}(b). 
It is observed that it reaches its maximum value at a proportion of 0.5. 
This suggests that having an equal proportion of users in both clients leads to the best performance. 
One possible reason for the increasing $F1$ score in testing features is that the algorithms require enough features to train a general model. 
Therefore, a higher proportion of features helps improve the performance. 
However, if the proportion becomes too large, the $F1$ score drops, indicating that having too many features to predict can hinder the model's performance.
The results demonstrate that if both clients have a similar number of users, the algorithm performs the best. 
This finding emphasizes the importance of achieving a balanced distribution of users across clients to optimize the model performance in federated learning scenarios.

\subsection{Discussion and Future Directions}
\noindent
The primary objective of this research was to address privacy concerns associated with sharing user data across organizations. 
The proposed algorithm successfully eliminates the need for user matching while still utilizing all features, thereby significantly enhancing user privacy. 
However, there are several aspects that should be further investigated and improved to advance the research in this area.
Firstly, it is important to conduct additional experiments to assess whether the sharing of feature vectors still indirectly discloses certain information about users. 
While the proposed algorithm eliminates the need for explicit user matching, future research should be conducted on developing advanced privacy-preserving techniques to minimize the disclosure of user information during the sharing of feature vectors. 
Another direction is to evaluate and generalize the proposed algorithm beyond the MovieLens dataset. 
It is important to assess the algorithm's performance on real-world datasets, for evaluating the algorithm's effectiveness and robustness across various domains and datasets.
Furthermore, expanding the experiments to involve three or more clients would be beneficial. 
This would provide insights into the algorithm's performance as the number of participating clients increases, as well as its scalability and effectiveness in real-world scenarios.

\section{Conclusion}
\label{sec:conclusion}
\noindent  
This paper presents FedDMF, an algorithm that tackles user attribute prediction while considering privacy concerns and legal requirements. FedDMF eliminates the need for user matching, safeguarding user privacy and preventing unauthorized access to sensitive information. By training deep matrix factorization models on individual clients and sharing only feature vectors, accurate user attribute prediction is achieved without compromising privacy. Experimental results on the MovieLens dataset show that FedDMF performs comparably to FedAvg, achieving 96\% of the accuracy of a single model. The proposed algorithm enhances customer targeting and improves the overall customer experience. FedDMF combines federated learning and deep matrix factorization, providing a promising solution for privacy-preserving user attribute prediction. It enables organizations to collaborate and leverage collective data while maintaining privacy and compliance. Future research should evaluate the algorithm on diverse datasets and explore optimization techniques for scalability and efficiency. FedDMF offers a practical and effective approach for leveraging user data while respecting privacy concerns.

\bibliographystyle{IEEEtran}
\bibliography{fl_c1}
\end{document}